\documentclass[runningheads]{llncs}
 
\usepackage[]{eccv}

\usepackage{eccvabbrv}
\usepackage{color}
\usepackage{graphicx}
\usepackage{booktabs}
\usepackage{amssymb,eucal}
\usepackage[dvipsnames]{xcolor}
\definecolor{mask_color}{RGB}{251, 231, 153}
\definecolor{gaussian_color}{RGB}{201, 223, 180}
\definecolor{rgb_color}{RGB}{194, 214, 238}

\usepackage{amsmath}

\usepackage{multirow}

\def\P{{ {\rm P} }}
\def\Loss{{\ell}}
\def\I{{\bf I}}

\usepackage[pagebackref,breaklinks,colorlinks,citecolor=eccvblue]{hyperref}

\usepackage{orcidlink}

\begin{document}

\title{Zippo: \underline{Zipp}ing C\underline{o}lor and Transparency Distributions into a Single Diffusion Model}

\titlerunning{Zipping Color and Transparency Distributions into a Single Diffusion Model}

\author{Kangyang Xie$^1$, ~ 
Binbin Yang$^4$, ~ Hao Chen$^1$, ~
Meng Wang$^2$ \\ Cheng Zou$^2$, ~ Hui Xue$^3$, ~ Ming Yang$^2$, ~ Chunhua Shen$^1$}

\authorrunning{Xie \textit{et al.}}

%
%

\institute{$^1$Zhejiang University 
~
$^2$Ant Group
~
$^3$Alibaba Inc.
~
$^4$Sun Yat-Sen University
}

\maketitle

\begin{abstract}
Beyond the superiority of the text-to-image diffusion model in generating high-quality images, recent studies have attempted to uncover its potential for adapting the learned semantic knowledge to
visual perception 
tasks. In this work, instead of irreversibly translating a generative diffusion model into a visual 
perception 
model, we explore to retain the generative ability 
with 
the perceptive adaptation.
To accomplish this, we present Zippo, a unified framework for \underline{zipp}ing the c\underline{o}lor and transparency distributions into a single diffusion model by 
expanding 
the diffusion latent into a joint representation of RGB images and alpha mattes. By alternatively selecting one modality as the condition and then applying the diffusion process to the counterpart modality, Zippo is capable of generating RGB images from alpha mattes and predicting transparency from input images. In addition to single-modality prediction, we propose a modality-aware noise reassignment strategy to further empower  Zippo with jointly generating RGB images and its corresponding alpha mattes under the text guidance.
Our experiments showcase Zippo's 
ability of efficient text-conditioned transparent image generation and present plausible results of Matte-to-RGB and RGB-to-Matte translation.

\end{abstract}

\begin{figure*}[ht!]
\begin{center}
\includegraphics[width=0.928\linewidth]{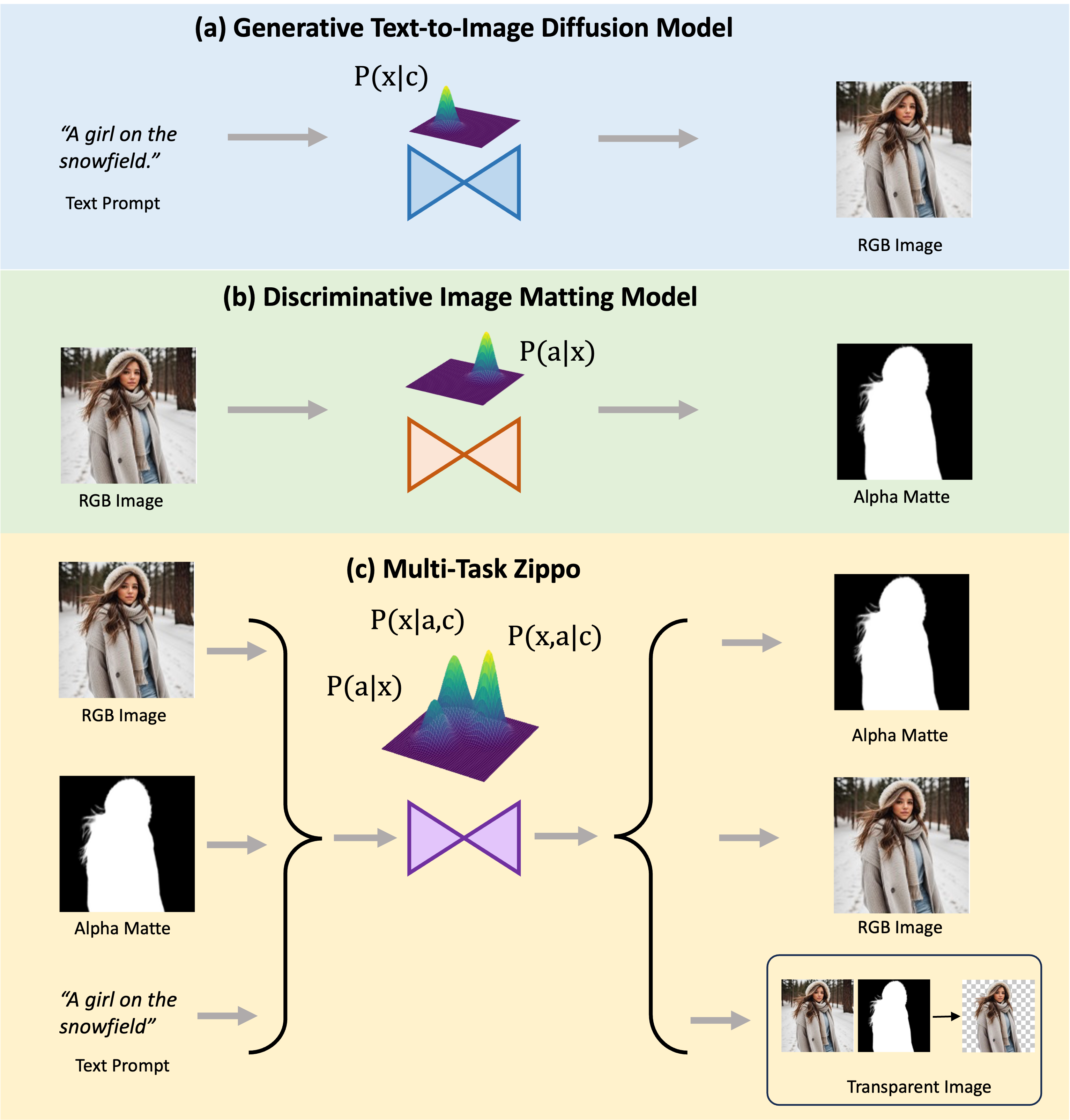}
\end{center}
\caption{An illustration of comparison between our proposed Zippo and a separate text-to-image diffusion model and image matting model. (a) text-to-image diffusion model generates a RGB image acccording to the input text prompt; (b) image matting model predict alpha matte from the input RGB image; (c) Zippo actually models the joint distribution of RGB and transparency information, thus acts as a multi-task learner. Specifically, besides the capabilities of RGB-to-Matte and Matte-to-RGB, Zippo can generate a pair of RGB image and alpha matte simultaneously, thus compositing a transparent image.}
\label{fig:fig1}
\end{figure*}

\section{Introduction}
\label{sec:intro}
The research of image generation has witnessed a rapid development in the past few months \cite{peebles2023scalable, podell2023sdxl, rombach2022highresolution}, thanks to the introduction of diffusion models. Their stunning generative performance implies that text-to-image diffusion models trained on massive data succeed to holistically capture the pixel affinities in the image, which are essential for synthesizing the coexisting objects within a scene \cite{chen2024deconstructing}. This observation motivates 
researchers 
to excavate the knowledge encapsulated within the generative diffusion model for downstream visual perceptive tasks \cite{ke2023repurposing, ji2023ddp, long2023wonder3d, xu2023openvocabulary, suzgun2024metaprompting,qi2023unigs,vangansbeke2024simple}. 
Image matting (\textit{a.k.a.}\ alpha matte prediction) is one of the important vision tasks, 
which aims to estimate the foreground's transparency of an input image.

Unfortunately, existing matting methods often suffer from scare data labels, thus resulting in poor generalization \cite{zhong2023lightweight, dim}. In this work, we aim to harness the embedded prior knowledge in a pre-trained text-to-image diffusion model, \emph{e.g.}, Stable Diffusion to derive a generalizable image matting model. 
Concretely, we propose a simple yet effective adaptation mechanism to finetune the diffusion UNet by inflating the UNet's input channel to encode the input image and perform diffusion process over the alpha matte.

However, directly inflating the input channel and translating the UNet into a matte predictor poses a significant problem. After tuning on the matte latent space, the diffusion UNet will gradually forget the seminal and versatile prior, and such irreversible adaptation will make it 
not be able to 
generate the RGB images from text prompts \cite{vangansbeke2024simple, ke2023repurposing, ji2023ddp}.
To solve this problem, we propose to incorporate another branch of generating RGB images from alpha mattes into the modified UNet. Symmetrical to the matting branch, we select the alpha matte latent $z_t^a$ as the condition while performing the diffusion process over the RGB latent $z_t^x$. Besides, to specify the routed branch, we introduce a modality indicator by learning a distribution identifier and inject the modality information into the diffusion UNet. By sharing the same UNet backbone between the two modalities, we can kill two birds with one stone, by zipping the color and transparency distributions into the single diffusion backbone. Such 
a design 
enables 
achieving both RGB-to-matte and matte-to-RGB within one model.

The 
proposed modality-aware noise reassignment strategy 
simply reassigns the noise sampled from two branches mentioned above to supervise the noisy joint latent $z_t = \texttt{cat}(z_t^x, z_t^a)$.
Zippo is empowered with learning one more joint distribution $\P(x,a|c)$ which means generating the RGB image and alpha matte simultaneously (namely transparent image) under the text guidance $c$. 

Thus, 
as illustrated in Fig.  \ref{fig:fig1}, we can 
view 
Zippo as a multi-modal distribution learner, involving $\P(x|a,c)$, $\P(a|x)$ and $\P(x,a|c)$ among color $x$, transparency $a$ and text prompt $c$. By contrast, the vanilla  T2I generative model \cite{zhang2023adding, rombach2022highresolution, saharia2022photorealistic} and discriminative matting model \cite{ma2023rethinking, chen2022ppmatting, yao2023vitmatte} in Fig.  \ref{fig:fig1} only partially captures $\P(x|c)$ or $\P(a|x)$, respectively. Therefore, Zippo actually acts as a multi-task learner which unifies the transparency-aware discriminative and generative paradigms.

To validate the effectiveness of our proposed framework, 
\textbf{Zippo}, we conduct extensive experiments on the datasets of AM2K \cite{li2021bridging} and VITON-HD \cite{choi2021vitonhd}. Results demonstrate that Zippo achieves accurate pixel-level semantic alignment between the RGB image and alpha matte, which is capable of accurately generating images from the given alpha mattes and predicting alpha masks from the given image. 
More importantly, by modeling the joint distribution, Zippo can simultaneously generate a pair of image and alpha matte, which equivalently predicts a transparent image. Such practice can be extensively applied for layered image composition \cite{lin2017layerbuilder}, facilitating professional digital image editing, texture mapping in computer graphics \cite{vinod2023teglo,chen2022AUVNET} and industrial film production  \cite{BMSengupta20}. For example, we can use Zippo to a generate transparent images and then composite 
foreground objects into a specific background arbitrarily through alpha blending \cite{vanaken2022alpha}. Without bells and whistles, our proposed Zippo has been demonstrated as a simple and flexible framework for generating both photo-realistic and editable transparent images.
\begin{figure*}[ht!]
\begin{center}
\includegraphics[width=1.0\linewidth]{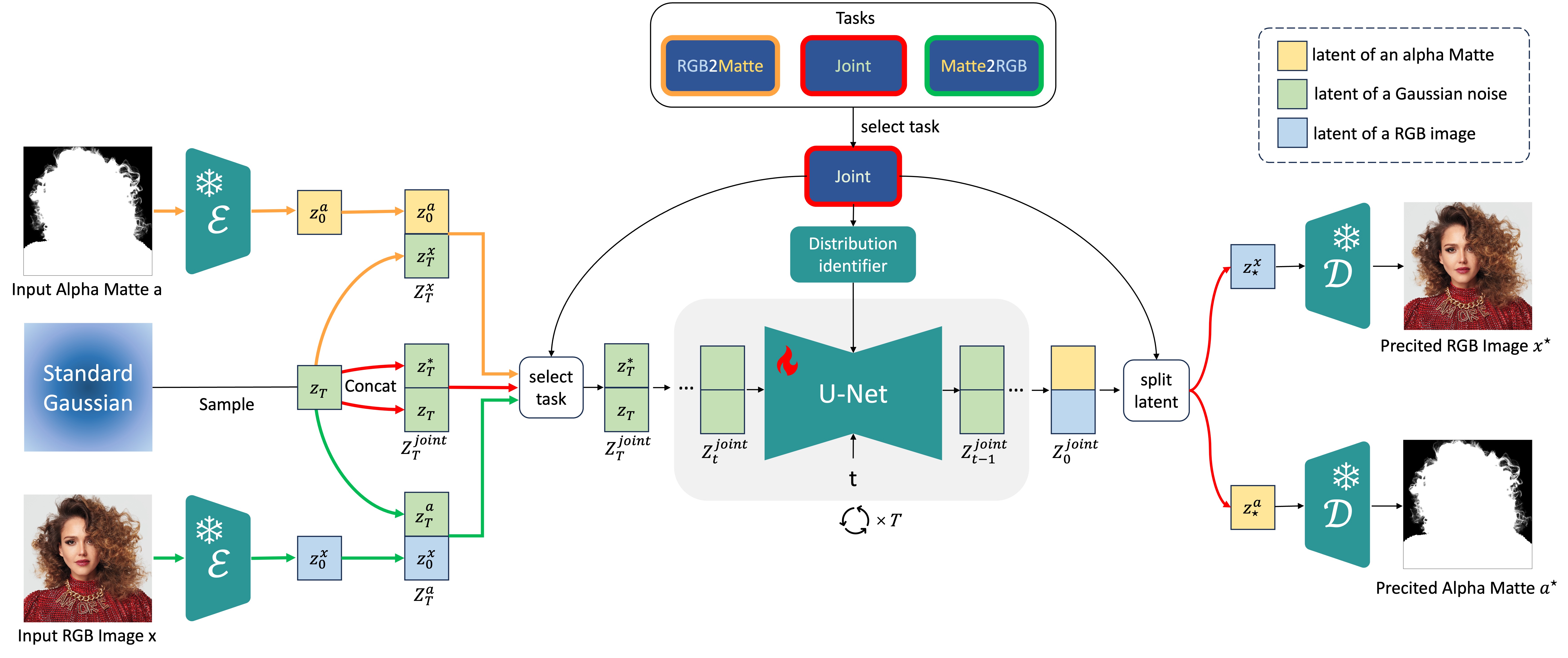}
\end{center}
\caption{An illustration of the workflow of our proposed Zippo. Given a RGB image $x$ and its corresponding alpha matte $a$, we turns the pre-trained generative diffusion model into a joint distribution learner of color and transparency information, which can perform the tasks of perceptive RGB-to-Matte estimation, Matte-to-RGB generation and jointly generating the paired image and alpha matte. For each task, we learn the distribution identifier for task routing. Taking the task of joint generation as example, we sample two standard Gaussian noises $z_T^*$ and $z_T$ and concatenate then together as the joint latent $z_T^{joint}$. Then Zippo iteratively denoise from noisy $z_T^{joint}$ to produce the clean joint latent $z_0^{joint}$. Finally, we split $z_0^{joint}$ to $z_*^x$ and $z_*^a$ and decode them into the RGB image $x^*$ and alpha matte $a^*$.
}
\label{fig:fig2}
\end{figure*}

\section{Method}
\label{sec:method}

\newcommand{\normaldistribution}{$N\left(0,I\right)$}

In this section, we first give a brief introduction to diffusion models in \S\ref{sec: preliminary}, then we formulate our problem in \S\ref{sec: formulation} as extending T2I diffusion models to generate images and corresponding alpha masks simultaneously, which exactly learns a joint distribution of color and transparency information. To this end, in \S\ref{sec: single dist modeling} we first propose to adapt the pre-trained T2I diffusion model to the task of alpha mask estimation. 
Additionally, we introduce another branch of Matte-to-RGB in \S\ref{sec: mask to rgb} to incorporate the generative colorful semantics.
Based on these two branches, we present a detailed discussion of our proposed model Zippo in \S\ref{sec: joint dist modeling}, which is posed to zip color and transparency distributions into single diffusion model that generates perceptually realistic images and high-quality alpha mask simultaneously.

\subsection{Preliminaries} 
\label{sec: preliminary}

In computer vision, diffusion models are first proposed to learn 
pixel RGB 
distribution by gradually predicting the noise in the perturbed data through a specifically designed parameterized reverse process \cite{ho2020denoising}. Concretely, the forward process is
that 
given an image $x_0$ drawn from the color distribution $\P(x)$,  a small amount of Gaussian noise $\epsilon$ is added to the image in $T$ steps, generating a sequence of noised images $x_1, x_2, \dots, x_T$ as:
\begin{equation}
\label{equation:add_noise}
    x_t = \sqrt{\bar{\alpha_{t}}}x_0 + \sqrt{1-\bar{\alpha_{t}}}\epsilon
\end{equation}
where $\epsilon \sim {\cal N}\left(0, {\bf I}\right)$ is standard normal distribution and $\bar{\alpha_{t}}=\prod_{i=1}^{t}1-\beta_{t}$, $\beta_t$ is controlled by the variance scheduler $\left\{\beta_{t} \in \left(0,1\right)\right\}_{t=1}^{T}$. As $t$ starts from 0 and gradually approaches $t=T$, the image $x_0$ is gradually noised and converges to $\epsilon \sim {\cal N}\left(0, {\bf I} \right)$. In the reverse process, the denoising model $\epsilon_{\theta}$ gradually removes noise by predicting the noise added to $x_{t-1}$ from $x_{t}$. 

To reduce the unaffordable training cost of generating high-resolution images, a more efficient diffusion training scheme is to perform the diffusion and denoising processes in the VAE latent space with encoder \cite{rombach2022highresolution} $\mathcal{E}$ and decoder $\mathcal{E}$:
\begin{equation}
    x_0 \approx \mathcal{D}(\mathcal{E}(x_0)).
\end{equation}

Further, by introducing the guidance \cite{ho2022classifierfree} of text prompt $c$, the learning object of a latent text-to-image diffusion model (Stable Diffusion) can be reduced to:
\begin{equation}
    \mathbb{E}_{t, \epsilon \sim {\cal N}(0, {\bf I})} \left[ \Vert \epsilon - \epsilon_\theta(x_t, t, c) \Vert_2^2\right].
\end{equation}

\subsection{Problem Formulation}\label{sec: formulation}
In this work, we aim to simultaneously zip the color and transparency information into single diffusion model. In other words, our diffusion UNet encodes adequate semantic information for (1) predicting mattes from RGB images; (2) generating RGB images from mattes; (3) simultaneously generating paired RGB images and mattes by the control of text prompts. 
Next, we 
formalize the learning paradigms for these three tasks: 
\begin{itemize}
    \item 

Given an image $x\in \mathbb{R}^{H\times W\times3}$, the alpha matte is predicted by learning the conditional distribution $\P(a|x)$; 
\item 
Given an alpha matte $a\in \mathbb{R}^{H\times W}$ and a text prompt $c$, the RGB image $x$ is generated by learning another conditional distribution $\P(x|a, c)$, where the text prompt $c$ is used to specify the colorful texture information in the image;
\item 
Given a text prompt $c$, the joint generation of paired image and alpha matte implies learning the conditional joint distribution $\P(x, a| c)$, where $c$ describes the RGB content to synthesize while $a$ represents the alpha transparency of the foreground in image $x$.
\end{itemize}

\subsection{Unified Latent Space for RGB and Alpha Matte}
\label{sec: unified vae}

\begin{figure*}[ht!]
\begin{center}
\includegraphics[width=1.0\linewidth]{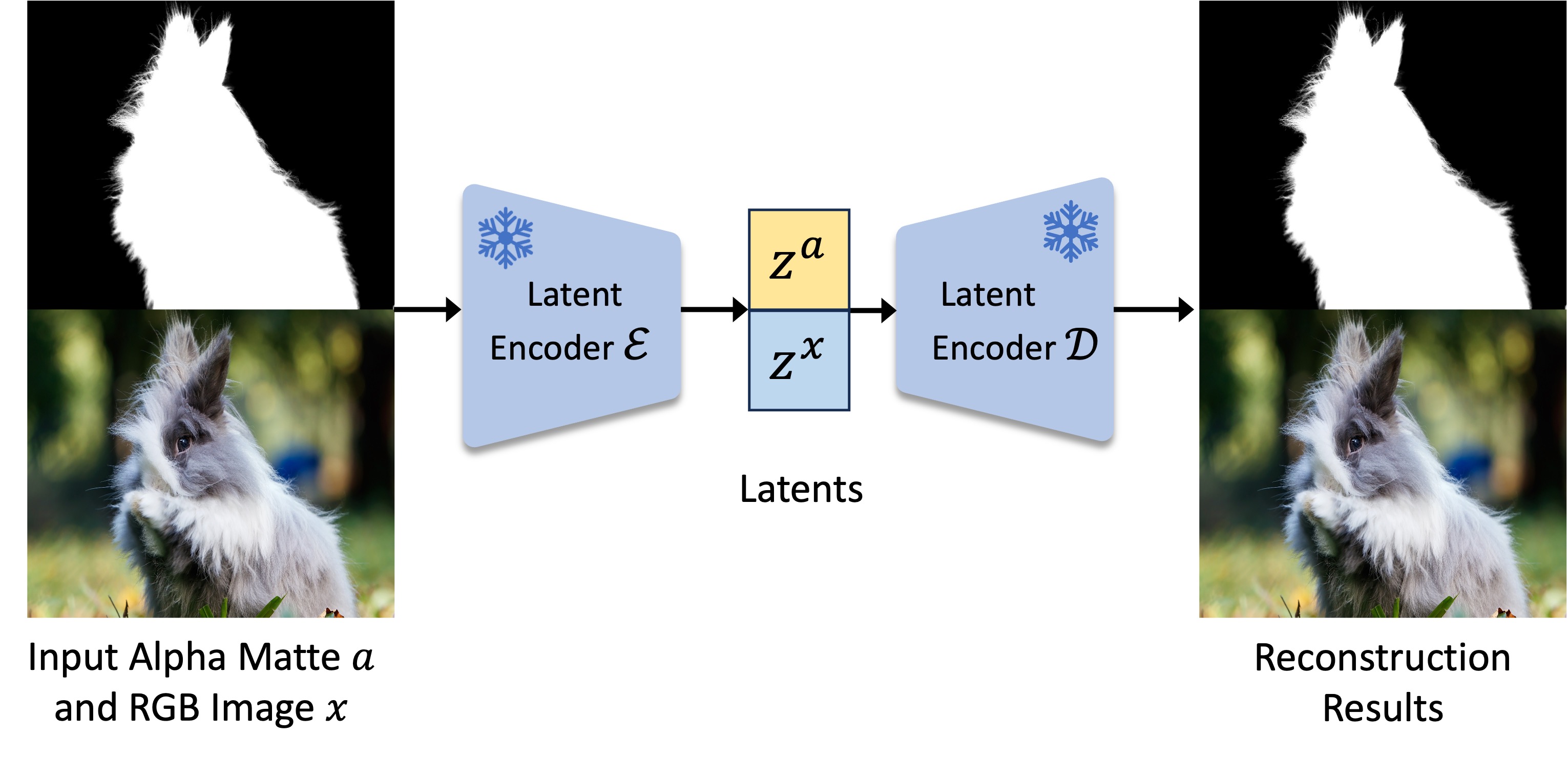}
\end{center}
\caption{RGB and Alpha matte share the same VAE encoder and decoder. A VAE trained for encoding RGB images is also sufficient for transparency reconstruction.}
\label{fig:vae}
\end{figure*}

According to the definition of alpha matte, unlike natural images, an alpha matte represents the transparency information through only single channel. it raises a question about how can we encode the alpha matte into the diffusion latent space.
Empirically, we 
find that 
a VAE \cite{kingma2022autoencoding} trained for encoding RGB images is sufficient 
to reconstruct the alpha matte by replicating the single-channel map into three channels. \begin{equation}
    x_0 \approx \mathcal{D}(\mathcal{E}(x_0)).
\end{equation}
Thus, 
the alpha matte and natural image can be encoded in the unified latent representation space: $z^a = \mathcal{E}(a)$ and $z^x = \mathcal{E}(x)$. As shown in Fig. \ref{fig:vae}, although it is not directly trained on the single-channel data, the pretrained VAE is also competent to robustly reconstruct the alpha matte with imperceptible differences.
As such, it is 
possible to share the same diffusion process between the RGB image and alpha matte.

\subsection{RGB-to-Matte Modeling} \label{sec: single dist modeling}
We first propose to adapt T2I diffusion model, \emph{e.g.}, Stable Diffusion to the alpha mask estimation task, which learns a conditional distribution $\P(a|x)$ over alpha masks $a \in \mathbb{R}^{H \times W}$ given an image $x \in \mathbb{R}^{H\times W\times 3}$. To incorporate conditional information into the diffusion UNet $\epsilon_\theta$, we apply the noising scheduler in Equation \eqref{equation:add_noise} to obtain the noisy matte latent $z_t^a$ and then concatenate the image latent $z^x$ with $z_t^a$, \emph{i.e.}, $z_t = \texttt{cat}(z^x, z_t^a)$ as the UNet's input. Due to the channel inflation of input latent, we duplicate the input channels of the $\mathtt{Conv_{in}}$ layer in the UNet. 

Then we perform the progressive denoising process \cite{song2022denoising} over the second half of the joint latent $z_t = \texttt{cat}(z^x, z_t^a)$, \emph{i.e.}, $z_t^a$ and obtain the predicted matte latent $z^a_*$. The training objective according to the RGB-to-Matte (R2M) task can be 
written 
as:
\begin{equation}
\Loss_\mathrm{R2M} = \mathbb{E}_{t, \epsilon \sim {\cal N}(0, \I)} \left[ \Vert \epsilon - \epsilon_\theta(z_t, t, \emptyset) \Vert_2^2\right],
\label{eq:training_objective_simple}
\end{equation}
where $\emptyset$ indicates that we use an empty prompt for such a perceptive task without content guidance.
Additionally, according to the aforementioned reconstructable compression of VAE in \S\ref{sec: unified vae}, the predicted matte $a^*$ for input image $x$ can be decoded by using the VAE decoder:
\begin{equation}
    a^* = \mathcal{D}(z^a_*).
\end{equation}

\subsection{Supplementing Matte-to-RGB Modeling}
\label{sec: mask to rgb}

 Although we have translated a generative T2I diffusion model into a visual perceptive model that estimates an alpha mask given an input image, sacrificing color image generation capability is 
 undesirable. 
 In this section, our goal is to retain the RGB image generation capability of the original pretrained T2I diffusion prior. To this end, we propose to supplement our translated RGB-to-Matte model (\S\ref{sec: single dist modeling}) with a complementary Matte-to-RGB branch.

As an opposite
problem to alpha mask estimation, generating an image from an alpha matte target at rendering an RGB image $x$ according to the alpha matte $a$ with the image content described by text prompt $c$, capturing the distribution $\P(x|a, c)$.
Different from existing works, we aim to retain the generative ability when translating the T2I diffusion model into an alpha matte estimator, which can be freely activated by a routing mechanism. To achieve this goal, the main challenges lie in (1) how to perform the multi-task learning; (2) how to decode the modality-specific information from the latent.

To solve these problems, we design a learnable distribution identifier, termed $d$ to specify the target distribution  (color/transparency). Concretely, the distributions involved are represented using one-hot vectors, and the distribution identifier $d$ is first encoded by a group of sine-cosine transformation and then added to the time-step embedding after a two-layer MLP  (Multi-layer Perceptron). 
Thus, 
we can perform multi-task learning and activate the routed branch by switching the distribution identifier to the target task distribution. While introducing the distribution identifier can achieve the task routing, it may cause confusion and conflict when encoding and decoding the modality-specific information. To remedy this issue, we replicate channel numbers of both first and last output layer of the UNet to independently encode and decode for different modalities. 
Then, given tripartite data $ (z^x,z^a,c)$, we can successfully extend the single-task perceptive diffusion model in \S\ref{sec: single dist modeling} into a dual-task distribution diffusion model $\epsilon_{\theta}^{**}\left(\cdot\right)$.

Formally, the training objective of the dual-task learner can be extended to:
\begin{equation}
\Loss_{dual} = \mathbb{E}_{\epsilon\sim {\cal N}(0, {\bf I});\hat{\epsilon}\sim {\cal N}(0, {\bf I});t}\left[ \left\| \epsilon -\epsilon_{\theta}(z^x, z^a_t, c, d)\right\|^2 + \left\| \hat{\epsilon} -\epsilon_{\theta}(z^x_t, z^a, c, d)\right\|^2\right]
\label{eq:training_objective_two_dist_simple}
\end{equation}

During inference time, our dual-task diffusion model is qualified for both RGB-to-Matte and Matte-to-RGB tasks. Given a target distribution identifier $d$, and a corresponding condition input (\emph{e.g.}, a color image $x$ or an alpha mask $a$) our dual-task diffusion model performs denoising diffusion process on the target noisy latent to estimate alpha matte or RGB image independently.

\subsection{Joint Modeling for Color  (RGB) and Transparency  (Alpha Matte)} 
\label{sec: joint dist modeling}

Although the dual-task paradigm proposed in \S\ref{sec: mask to rgb} could perform two branch modeling, it only separately models the color and transparency distribution $\P(x|a,c)$ and $\P(a|x)$. However, modeling the transparency-aware joint distribution, \emph{i.e.}, $\P(x, a | c)$ is a more meaningful and applicable task with immense industrial and commercial demands. For example, we can obtain the transparent image by alpha blending the RGB image $x$ and the alpha matte $a$, which holds significant practical value in the field of layered image composition, film production and texture mapping.

In this section, our goal is to go one step further: empowering the dual-task diffusion model introduced in \S\ref{sec: mask to rgb} with the capability of modeling the joint distribution $\P(x,a|c)$.
To this end, we present Zippo, a unified generative framework to zip color and transparency distributions into single diffusion model equipped with three branches architecture and modality-aware noise reassignment training strategy to model the joint distribution $\P(x,a|c)$ collaboratively with two conditional distribution $\P(x|a,c)$ and $\P(a|x)$.

In particular, 
we implement the training strategy by reusing the sampled noises ($\epsilon$ and $\hat{\epsilon}$) in \S\ref{sec: single dist modeling} and \S\ref{sec: mask to rgb} to perturb the matte and RGB latent  ($z^a$ and $z^x$). Then the UNet takes the joint latent $z_t = \mathtt{cat}(z^x_t, z^a_t)$ and the text prompt $c$ as input to predict the joint noise $\mathtt{cat}(\hat{\epsilon}, \epsilon)$. Training with the joint noise estimation enables Zippo to learn one more joint distribution $\P(x,a|c)$ and generate paired RGB image and matte under the guidance of text prompt $c$.

To achieve this, we reuse the sampled noises ($\epsilon$ and $\hat{\epsilon}$) in  \S\ref{sec: single dist modeling} and  \S\ref{sec: mask to rgb} to perturb the matte and RGB latent ($z^a$ and $z^x$). Then the UNet takes the joint latent $z_t = \mathtt{cat}(z^x_t, z^a_t)$ and the text prompt $c$ as input to predict the joint noise $\mathtt{cat}(\hat{\epsilon}, \epsilon)$. By training with the joint noise estimation, Zippo learns the joint distribution of color and transparency and is capable of generating paired RGB image and matte with the guidance of text prompt.

\section{Experiment}
\label{sec:exp}
\subsection{Experimental Setting}

\subsubsection{Datasets}
Zippo is trained on two datasets AM2K \cite{li2022bridging} and VITON-HD \cite{choi2021vitonhd}. Considering the important human-centric characteristics of generating transparent portrait images for vast applications  (\textit{e.g.}, film scene/character composition, professional portrait layer composition, and layout design), it is essential to gather high-quality unobstructed 
portrait images for training Zippo. While P3M10K \cite{li2021privacypreserving} is a commonly used dataset in portrait matting, we have opted not to utilize it due to the privacy masks presented on human faces, which could be harmful to the quality of portrait generation. Therefore, we use an open-source virtual try-on dataset VITON-HD, which has 10k high quality human-centered images without masks on the face. For high-quality alpha matte labels, we first choose trimap-free portrait matting model P3M \cite{li2021privacypreserving} to obtain the coarse alpha labels. Then we predict segmentation masks leveraging the strong semantic prior of SAM trained on 11B data \cite{kirillov2023segment} and filter out those that are deviated from the alpha labels. In concert, we quantize the alignments between alpha matte (binarized by a threshold $h=25$) and segmentation masks through computing IoU  metric, and we only collect alpha labels that have IoU  result greater than the threshold $\hat{h}$=0.9 \cite{chicherin2023adversariallyguided}. Finally, we estimate a trimap from the filtered alpha mask as an auxiliary input for trimap-based matting model \cite{park2022matteformer} which outputs the final fine-grained alpha labels. To explore the generalization of our model, we also adapt AM2K as another training data which has 2k high-quality animal images with human-labeled high-quality alpha mattes. 

\begin{figure}[htb!]
\begin{center}
\includegraphics[width=0.8\linewidth]{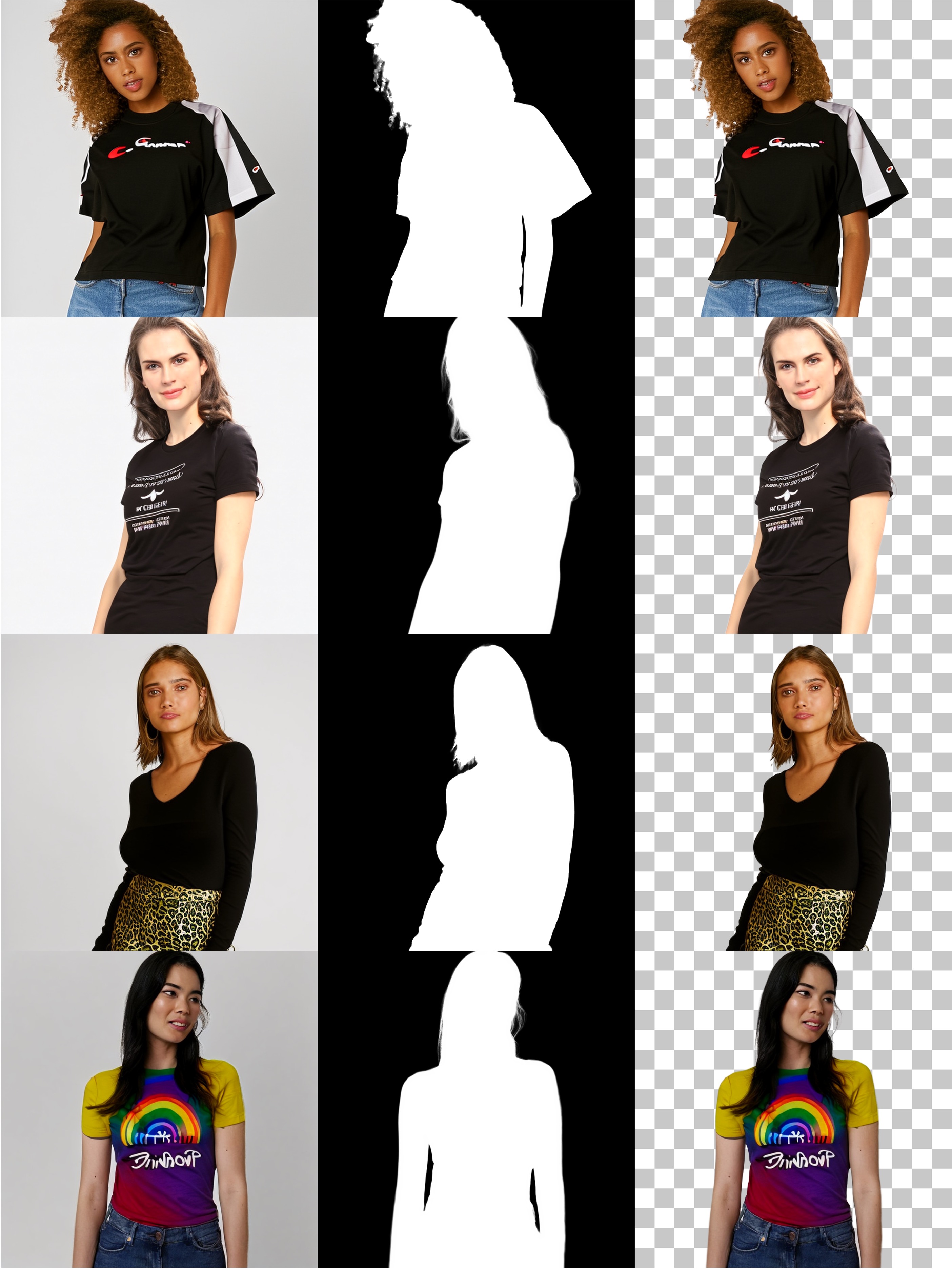}
\end{center}
\caption{Results of joint generation on VITON. The first two columns are generated simultaneously from Zippo. To evaluate the alpha mask, we present the result transparency image composed from the first two columns.
}
\label{fig:joint viton}
\end{figure}

\subsubsection{Implementation Details}
We build Zippo upon the publicly available Stable Diffusion v2.1 model to leverage the strong prior pretrained on the large-scale text-image dataset \cite{schuhmann2022laion5b}. During training, we fintune the whole UNet model initialized from the pretrained weights. 
For the training data organization, the batch data $batch$ containing paired images and alpha mattes are replicated three times and concat together on the batch dimension (default the first dimension) namely $ new\_batch = {\tt concat} (batch,batch,batch)$ for training three tasks simultaneously in one single forward pass. 
For hyper-parameters setting, we set lr as 1e-5 with cosine scheduler, a batch size of 6, cropping the image to resolution $768 \times 768$. 

\begin{figure*}[ht!]
\begin{center}
\includegraphics[width=0.8\linewidth]{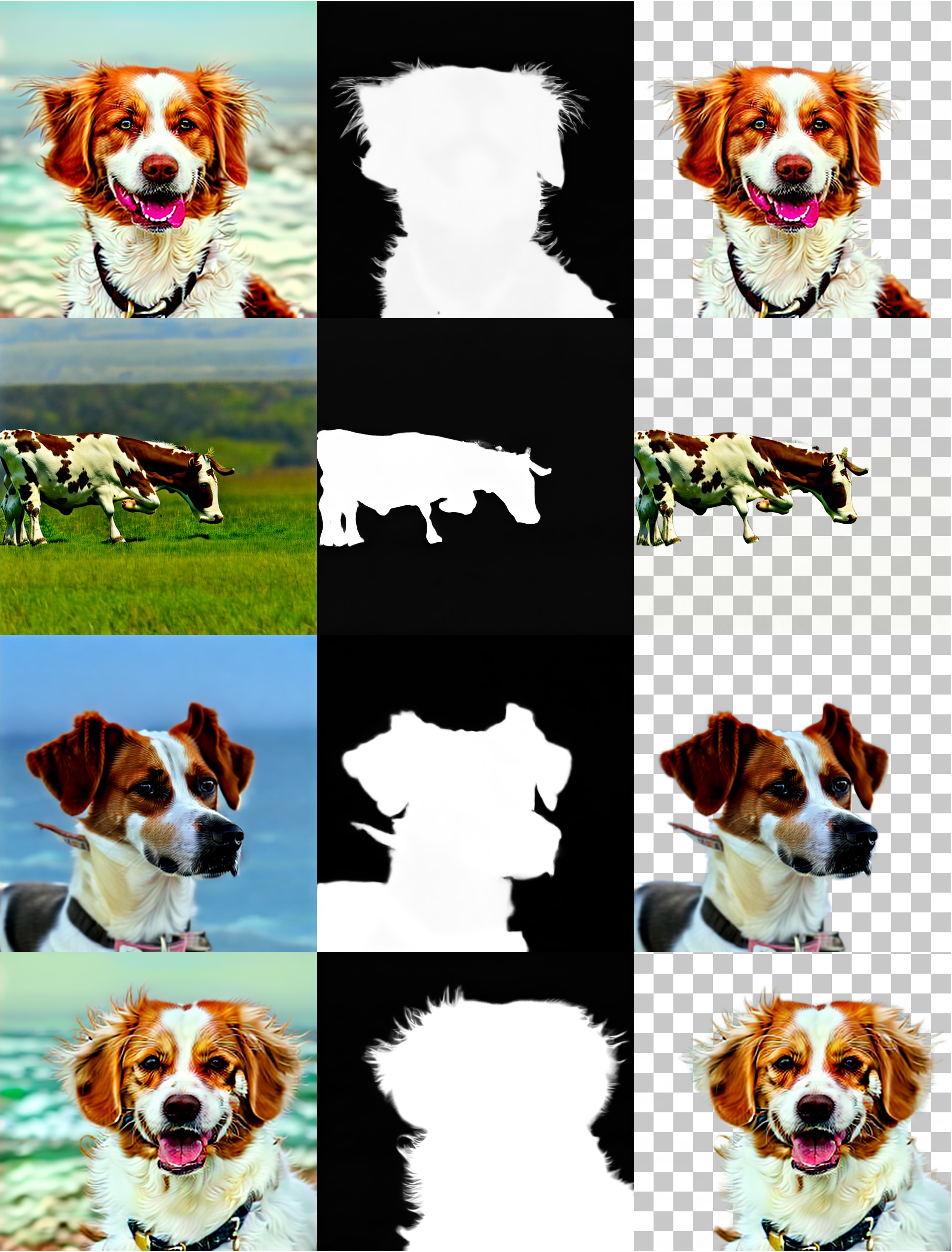}
\end{center}
\caption{Results of joint generation on AM2K. The first two columns are generated simultaneously from Zippo. To evaluate the alpha mask, we present the result transparency image composed from the first two column.
}
\label{fig:joint am2k}
\end{figure*}

\subsection{Main Results}
In this section, we present the results of Matte-to-RGB, RGB-to-Matte and joint generation (transparent image generation) on the datasets of AM2K and VITON. In Figs. \ref{fig:joint am2k} and \ref{fig:joint viton}, we present the results of joint generation of RGB images and corresponding alpha mattes. From the results, we can conclude that Zippo generalizes well to the cases of both animal and human. It is noteworthy that Zippo can not only generate photo-realistic RGB images but also simultaneously output its corresponding fine-grained alpha matte. By combining these two outputs together, we show the composed transparent images to verify the accurate alignment between RGB and matte and the content adherence to the text prompt. For example, in Fig. \ref{fig:joint am2k}, Zippo generated a furry dog with accurate reasonable details, even predicting alpha transparency value for each strand of dog fur.

\begin{figure*}[ht!]
\begin{center}
\vspace{-4mm}
\includegraphics[width=0.8\linewidth]{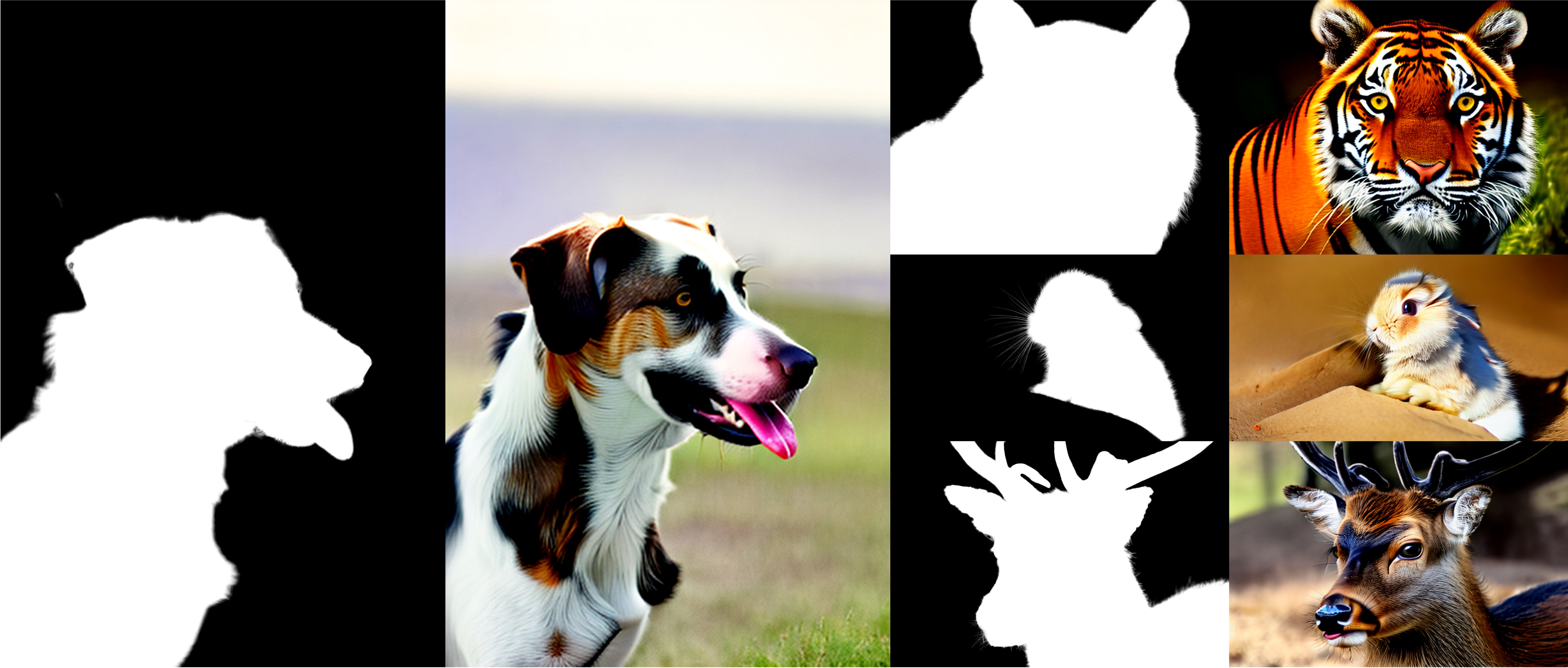}
\end{center}
\caption{Results of Matte-to-RGB generation on AM2K. The first and third columns are input alpha mattes and the second and the forth columns are the generated results by Zippo. The text prompts are "a dog", "a tiger", "a rabbit" and "a deer", respectively.
}
\label{fig:mask2rgb am2k}
\vspace{-3mm}
\end{figure*}

\begin{figure*}[ht!]
\begin{center}
\vspace{-4mm}
\includegraphics[width=0.8\linewidth]{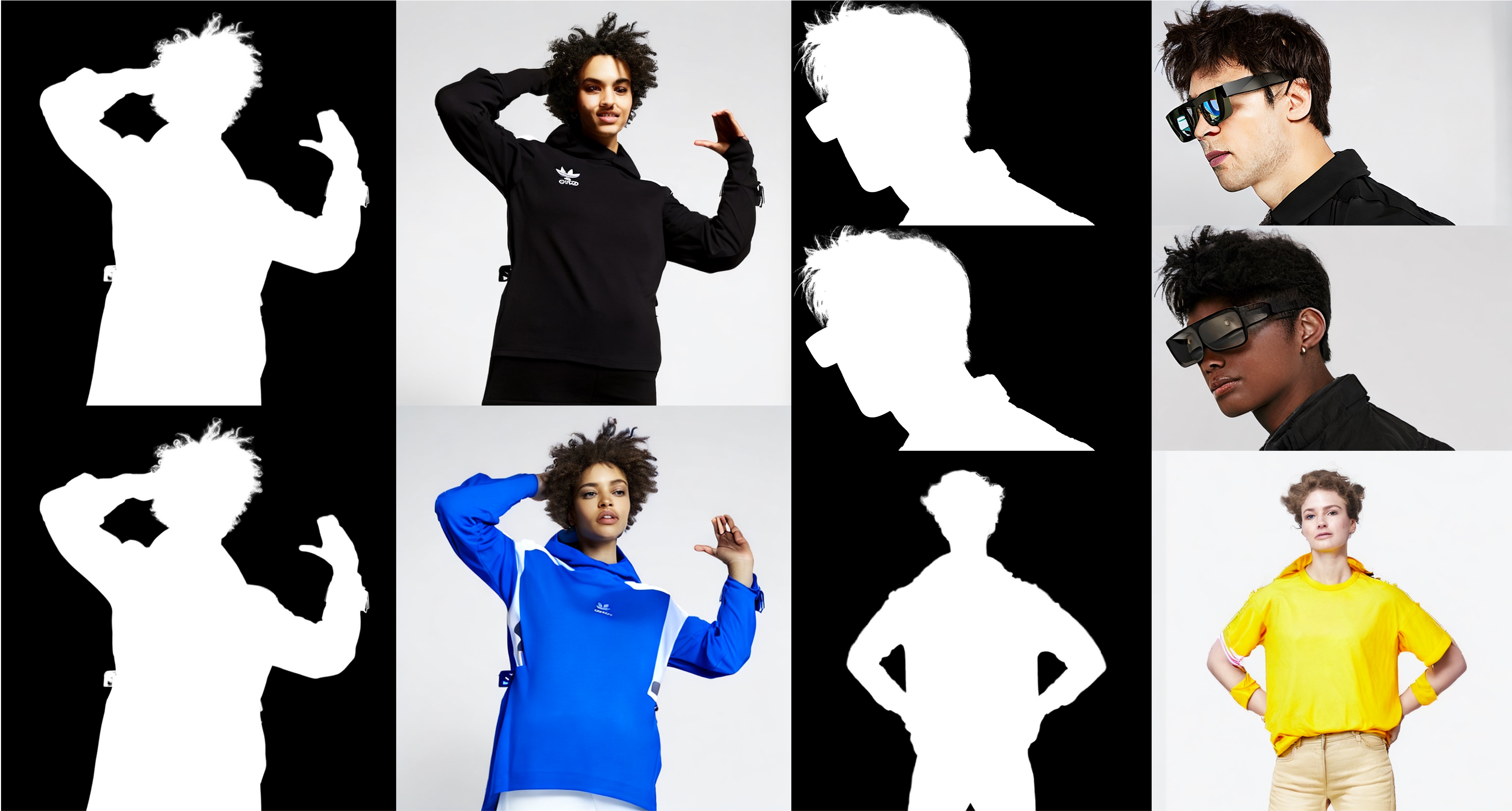}
\end{center}
\caption{Results of Matte-to-RGB generation on VITON. The first and third columns are the input alpha matte, and the second and fourth columns present the images in different resolutions generated by Zippo. Zoom in for the detailed alignments. For the first column, the text prompts are "a man wearing a black hoodie", and "a woman wearing a blue hoodie". For the third column, the text prompts are "a man", "a woman" and "a woman wearing a yellow t-shirt." 
}
\label{fig:mask2rgb viton}
\vspace{-3mm}
\end{figure*}

\begin{figure*}[ht!]
\begin{center}
\includegraphics[width=0.8\linewidth]{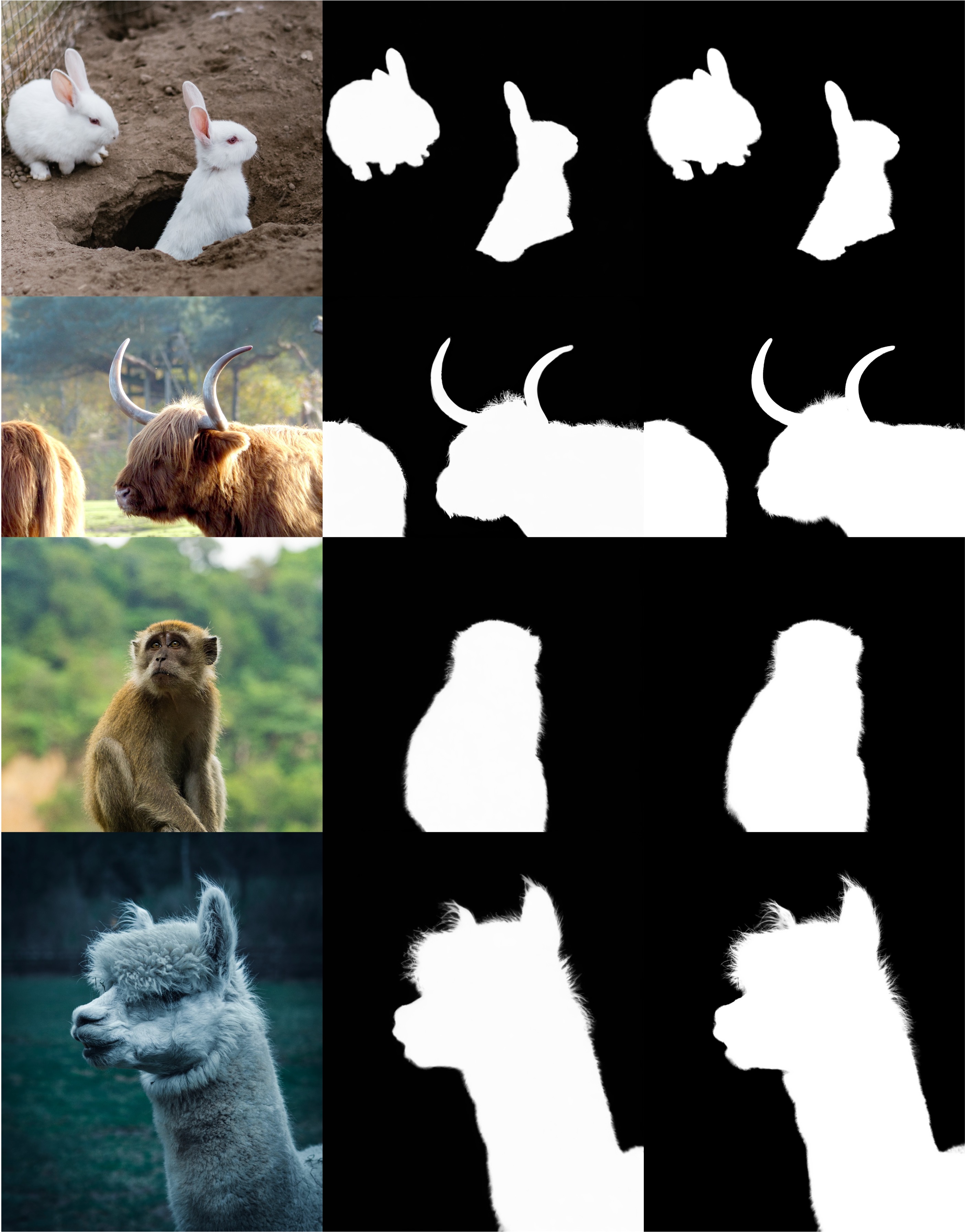}
\end{center}
\caption{Results of RGB-to-Matte prediction on AM2K. The first column shows input images in the test set, the second column shows predicted alpha mattes, and the third column shows the ground truth. It's worth noting that Zippo presents great details compared to the ground truth alpha matte.}
\label{fig:rgb2mask am2k}
\end{figure*}

\begin{figure*}[ht!]
\begin{center}
\includegraphics[width=0.8\linewidth]{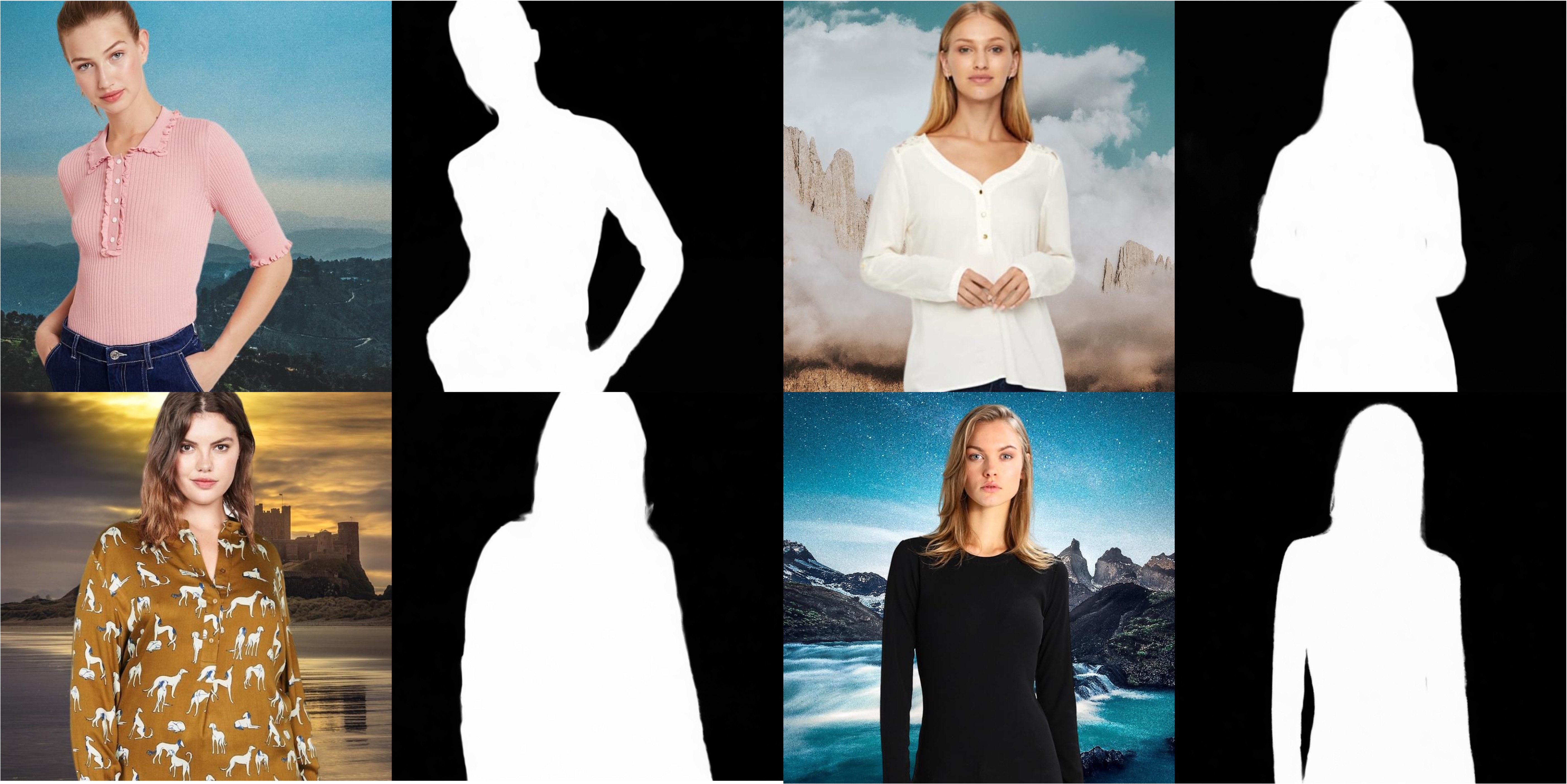}
\end{center}
\caption{Results of RGB-to-Matte prediction on VITON. The first and third column is the input image, and the second and fourth column is the predicted alpha matte.
}
\label{fig:rgb2mask viton}
\end{figure*}

Additionally, we provide the results of Matte-to-RGB and RGB-to-Matte on AM2K and VITON in Figs. \ref{fig:mask2rgb am2k}, \ref{fig:mask2rgb viton}, \ref{fig:rgb2mask am2k}, \ref{fig:rgb2mask viton}. In Figs.  \ref{fig:mask2rgb am2k} and \ref{fig:mask2rgb viton}, we take the alpha masks as the structural inputs and use the text prompts to specify the contents in the images, \emph{e.g.}, the classes of animals in the images and the clothes worn by the people. As can be concluded, our Zippo is capable of synthesizing animals and humans  (Matte-to-RGB) with structural adherence to the input alpha mattes and plausible photo-realistic contents with fancy details. In Figs. \ref{fig:rgb2mask am2k} and \ref{fig:rgb2mask viton}, we take the RGB images as input and use Zippo to predict the corresponding alpha mattes (RGB-to-Matte). As shown in Fig. \ref{fig:rgb2mask am2k}, Zippo is able to estimate the alpha mattes for input images of arbitrary resolutions and successfully mat the foreground objects with detailed masks.

\section{Related Work}

\noindent\textbf{Text-to-Image Diffusion Models} Text-to-Image diffusion models are a class of diffusion-based image generative models that aim to generate images with content semantically aligned with the given text prompt\cite{ho2020denoising}. Text-to-image diffusion models start from random Gaussian noises taking text prompt as the input and gradually denoise the noise under the guidance to output an image in high fidelity. Recently, many text-to-image diffusion models have been proposed by giant companies as their magical generation fidelity and promising future applications. Rombach et al. trained a diffusion model in latent space called Stable Diffusion\cite{rombach2022high} on a large-scale image and text dataset, LAION-5B, and demonstrated previously unachievable image synthesis quality. 

\noindent\textbf{Generative models for visual perceptive tasks}
Generative models capture strong images prior by training on large-scale text-image datasets. Recently, increasingly works explored to leverage the priors of generative models for visual perceptual tasks. Recent works~\cite{ji2023ddp, xu2023openvocabulary,long2023wonder3d} demonstrate that generative models are capable of solving dense prediction tasks like depth estimation, segmentation, and normal estimation. By finding latent variable offsets\cite{bhattad2023stylegan}, generative models can directly produce intrinsic images like surface normals, depth, albedo, etc. Marigold~\cite{ke2023repurposing} fine-tunes unet backbone of diffusion models to perform affine-invariant monocular depth estimation and exhibit strong generalization capability. 
Meta prompt~\cite{wan2023harnessing} proposes to harness generative models for visual perceptive tasks by introducing learnable embeddings to the pre-trained generative models to extract proper features for visual perceptive tasks. 

\noindent\textbf{Image matting.}
Image matting is a typical ill-posed problem to estimate the foreground's alpha matte from a single image. From the perspective of auxiliary input, image matting can be categorized into two types of methods. Trimap-based image matting methods require an auxiliary input trimap to mitigate the uncertainty\cite{yao2023vitmatte, xu2017deep ,park2022matteformer, lu2019indices}. Meanwhile, researchers attempt to learn an automatic end-to-end model without any auxiliary inputs namely, trimap-free image matting methods\cite{li2021privacypreserving, ke2022modnet,zhong2023lightweight, chen2022ppmatting}. Previous methods include affinity based methods\cite{aksoy2017designing,levin2007closed} and sampling-based methods\cite{he2011global, wang2007optimized}. Although trimap-free image matting methods are more applicable and flexible, they still fall behind the trimap-based matting methods according to their performance. Therefore, we design our pesudo-labeling roadmap by first obtaining the coarse alpha mask from trimap-based method, then computing a trimap according to the alpha mask and input it to the trimap-based method for a fine-grained alpha matte as our final pesudo-labels. It is worth noting that, previous image matting methods are typical distcriminative ones by regressing the estimation results, while we try to solve this problem in a generative manner, by formulating the image mattiing problem as learning a conditional alpha matte distribution given the input images. the input images.

\section{Conclusion}
In this paper, we propose a unified framework, 
termed 
Zippo, for zipping the joint color and transparency distributions into a single diffusion model. We first harness the learned versatile prior of pre-trained text-to-image diffusion model for task of the alpha matte estimation. Specifically, we propose to translate the generative stable diffusion model to a perceptive matting model by inflating the UNet input channels and fine-tuning the UNet. Besides, we explore to retain the generative ability after the perceptive adaptation and obtain a dual-task learner of RGB-to-Matte and Matte-to-RGB. Furthermore, we aim to learn the joint distribution of color and transparency in the same diffusion space to enable simultaneous generation of paired RGB image and the alpha matte, which can be used for compositing a transparent image. Last but not least, Our experiments showcase Zippo's extensive ability of efficient text-conditioned transparent image generation and present plausible results of Matte-to-RGB and RGB-to-Matte.

\clearpage

\section*{Appendix}

\appendix

\section{Quantitative Results of RGB-to-Matte}
We present the quantitative comparison of RGB-to-Matte with an image matting method, \emph{i.e.}, PPMatte and the ablation studies for our Zippo in Table \ref{tab: quantitative}. The results demonstrate that our proposed Zippo achieved remarkable matting performance on the AM2K dataset, even outperforming the matting-specific model PPMatte. Besides, we provide the quantitative results of three Zippo variants, \emph{i.e.}, $\text{Zippo}_\text{single}$, $\text{Zippo}_\text{two}$ and $\text{Zippo}_\text{full}$. To be specific, $\text{Zippo}_\text{single}$ means the simple extension of translating the T2I diffusion model into a matting model, $\text{Zippo}_\text{two}$ means the further extension of introducing the Matte-to-RGB branch, while $\text{Zippo}_\text{full}$ is our ultimate full model. As shown in Table \ref{tab: quantitative}, the RGB-to-Matte branch will benefits from incorporating the RGB generation and jointly generating RGB and Matte, due to the introduction of fine-grained pixel supervision.

\section{Controllable Content for Matte-to-RGB}
\begin{figure*}[ht!]
\begin{center}
\includegraphics[width=1.0\linewidth]{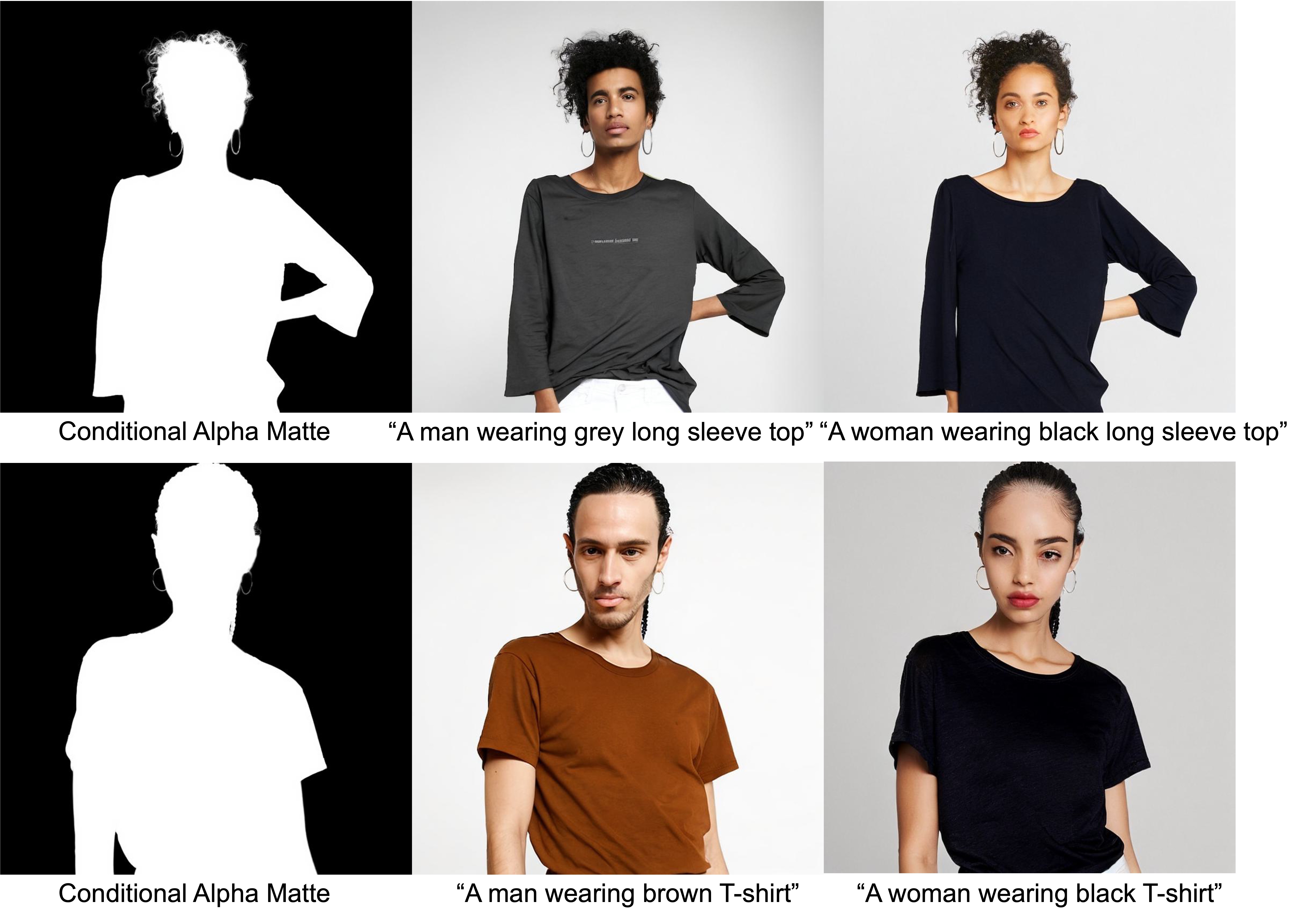}
\end{center}
\caption{Visualization of controllable content.}
\label{fig:control_content}
\end{figure*}

We provide the results of Matte-to-RGB generation in Fig. \ref{fig:control_content}, where the image content is both controlled by the conditional alpha matte and the text prompt. As shown in the figure, we can control the gender and the cloth of generated human in the image using the textual guidance. For example, although using the same matte, we can generate a man or a woman with the same hair style and the earrings.

\section{Comparison of Matte-to-RGB with ControlNet}
\begin{figure*}[ht!]
\begin{center}
\includegraphics[width=1.0\linewidth]{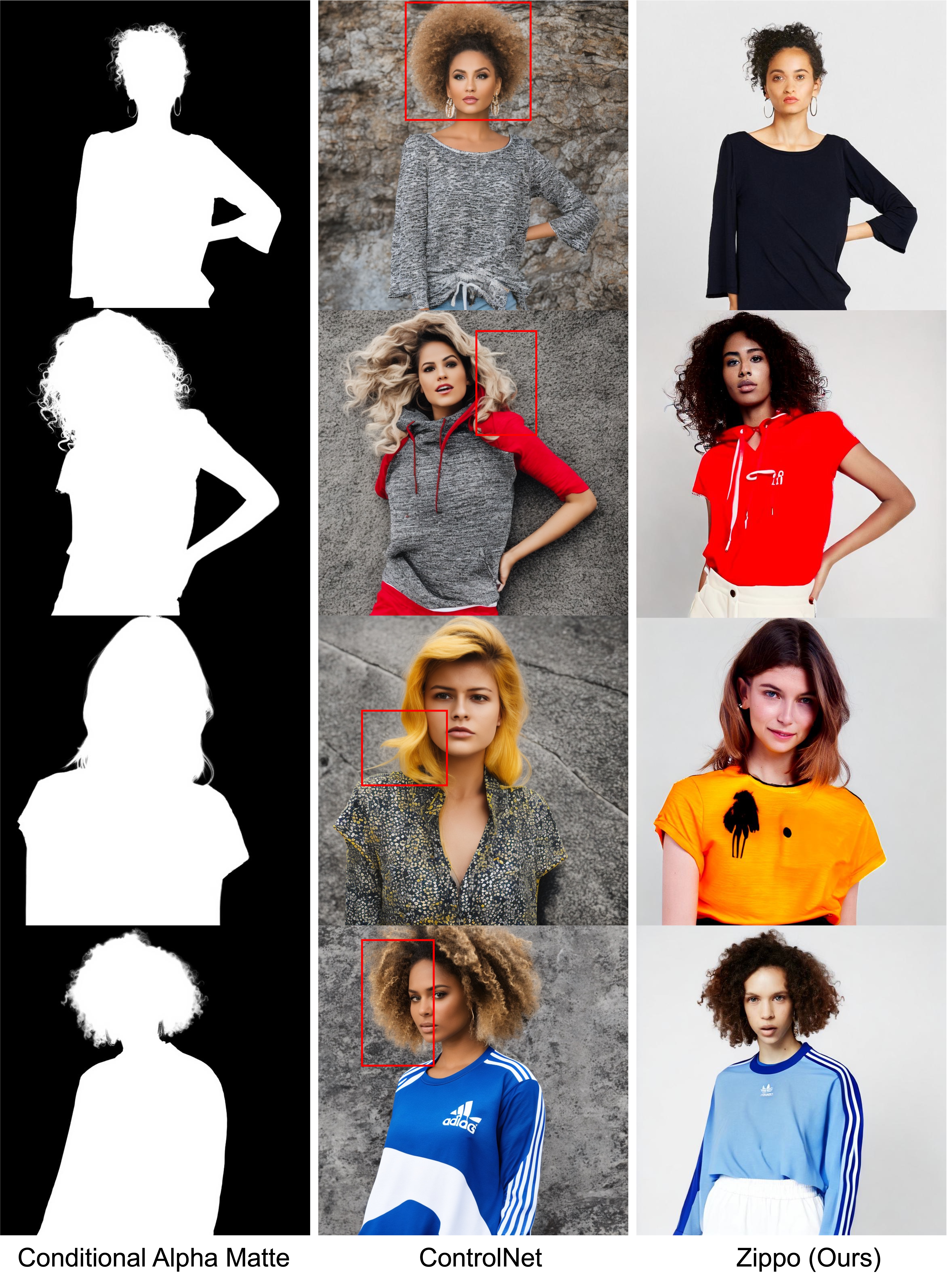}
\end{center}
\caption{Comparison to ControlNet.}
\label{fig:comp_controlnet}
\end{figure*}

To further evaluate the Matte-to-RGB performance of our Zippo, we provide the comparison with ControlNet in Fig. \ref{fig:comp_controlnet}. As shown in the result, although ControlNet provides a plausible results with coarse alignment with the conditional alpha matte, Zippo shows more accurate and fine-grained transparency control. For example, ControlNet fall short of aligning the hair details while Zippo presented appealing results.

\section{Comparison of Joint Generation with DiffuMask}
\begin{figure*}[ht!]
\begin{center}
\includegraphics[width=1.0\linewidth]{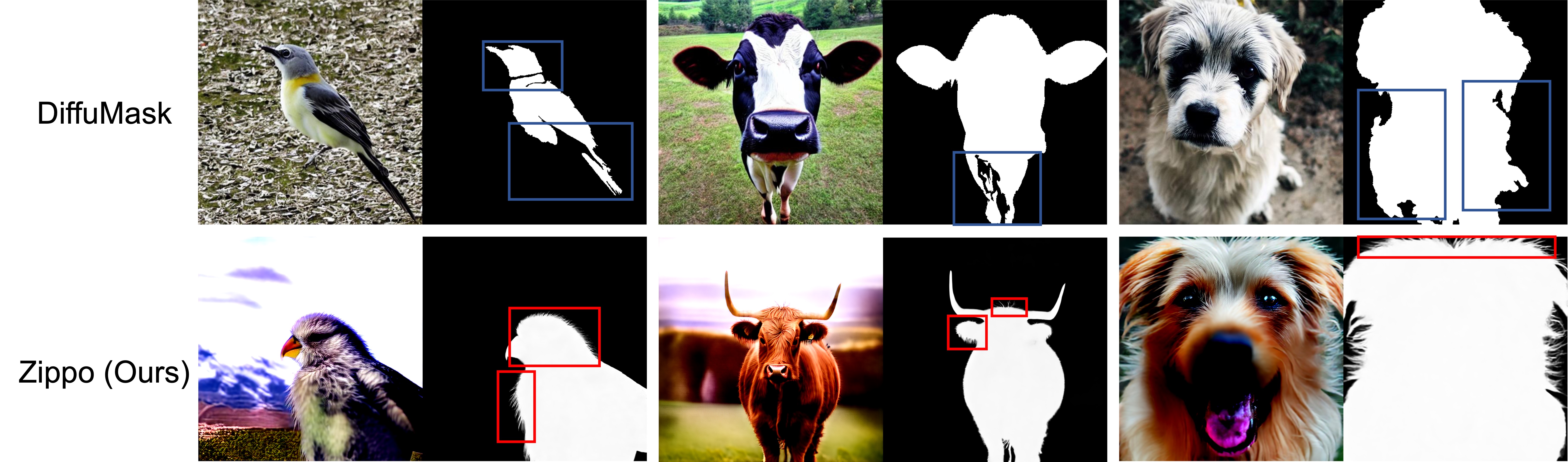}
\end{center}
\caption{Comparison to DiffuMask.}
\label{fig:comp_diffumask}
\end{figure*}

To demonstrate the performance of joint generation of matte and image, we compare Zippo to DiffuMask~\cite{wu2023diffumask} in Fig. \ref{fig:comp_diffumask}. The results show that DiffuMask only outputs coarse-grained mask fragments, and is incapable of generating detailed alpha masks. By contrast, Zippo can accurately generate images and the paired alpha mattes.

\begin{table}[h]
  \centering
  \caption{Quantitative comparisons
  }
\begin{tabular}{r|c|c|c|c|c|c}
    \toprule
    \multirow{2}{*}{Method} & \multicolumn{6}{c}{Metrics} \\
    \cline{2-7}
	 & SAD$\downarrow$ & MSE$\downarrow$ & MAD$\downarrow$ & $\text{SAD}_\text{FG}$$\downarrow$ & $\text{SAD}_\text{BG}$$\downarrow$ & CONN$\downarrow$ \\
    \hline
        PPMatte~\cite{chen2022pp} & 30.038 & 0.017 & 0.017 & 9.970 & 8.156 & 29.404 \\
        $\text{Zippo}_\text{single}$ & 23.449 & 0.006 & 0.0138 & 4.051 & 4.273 & 18.192 \\
        $\text{Zippo}_\text{two}$ & 22.973 & 0.005 & 0.014 & 3.712 & 5.638 & 15.999 \\
        $\text{Zippo}_\text{full}$ & \textbf{17.485} & \textbf{0.004} & \textbf{0.010} & \textbf{2.697} & \textbf{2.705} & \textbf{13.329}\\
    \bottomrule
  \end{tabular}
  \label{tab: quantitative}
\end{table}

\begin{figure*}[ht!]
\begin{center}
\includegraphics[width=1.0\linewidth]{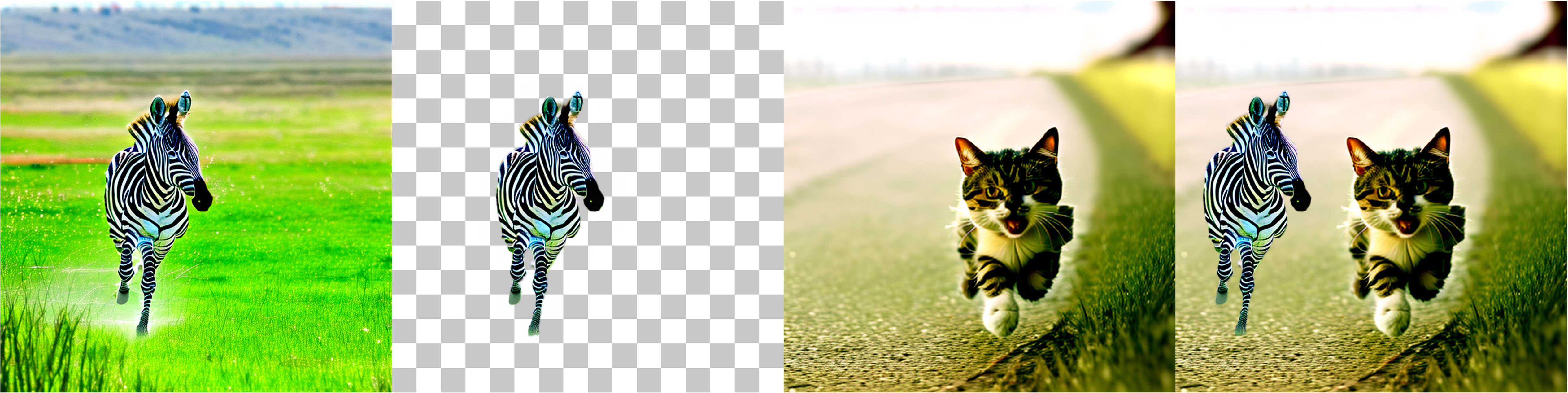}
\end{center}
\caption{Composition result. We compose two images together using the alpha matte generated simultaneously with the image. Note that both images are generated from Zippo.}
\label{fig:compose1}
\end{figure*}

\begin{figure*}[ht!]
\begin{center}
\includegraphics[width=1.0\linewidth]{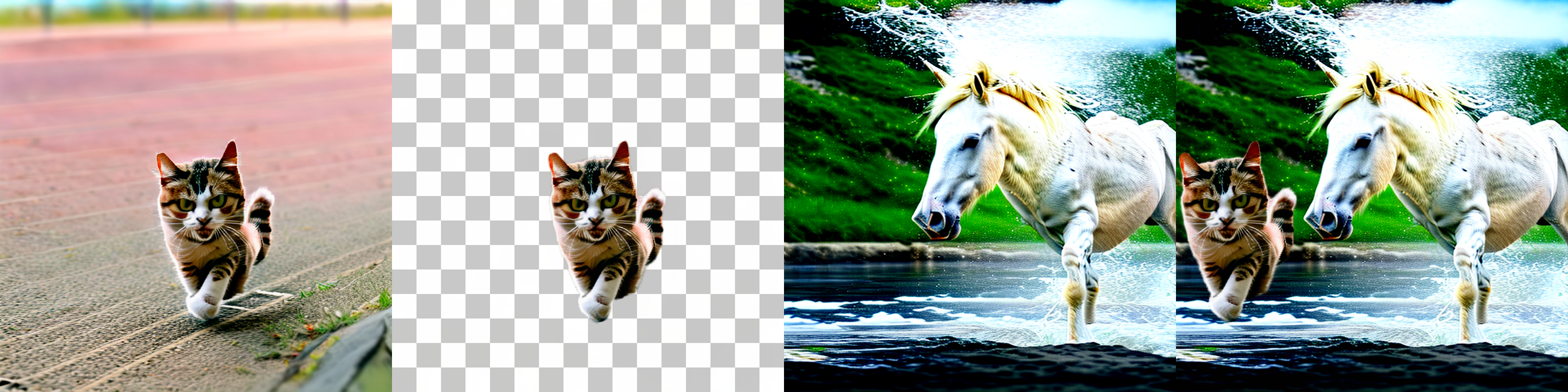}
\end{center}
\caption{Composition result. We compose two images together using the alpha matte generated simultaneously with the image. Note that both images are generated from Zippo.}
\label{fig:compose2}
\end{figure*}

\begin{figure*}[ht!]
\begin{center}
\includegraphics[width=1.0\linewidth]{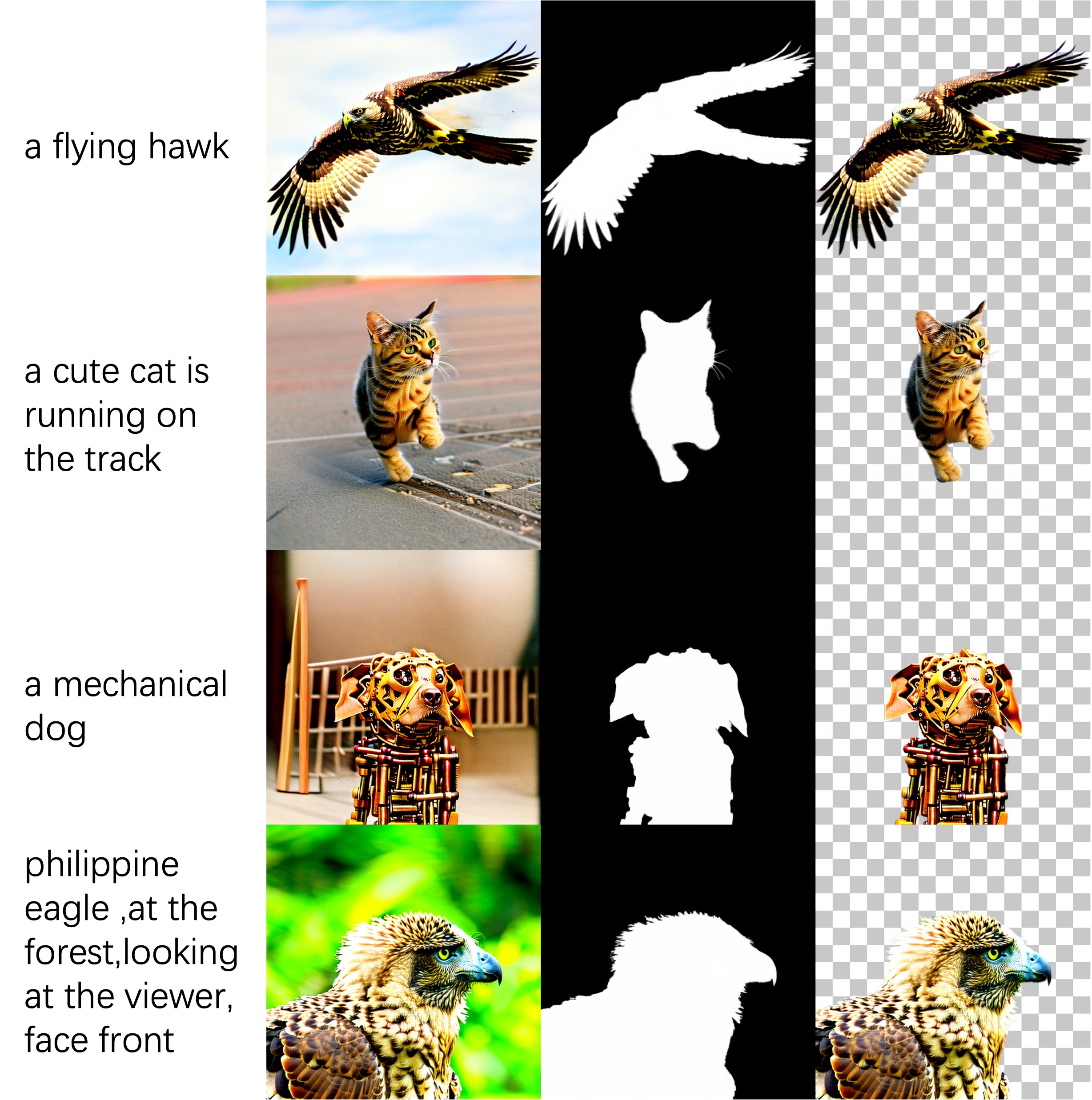}
\end{center}
\caption{More jointly generated results from Zippo.}
\label{fig:am2k1}
\end{figure*}

\begin{figure*}[ht!]
\begin{center}
\includegraphics[width=1.0\linewidth]{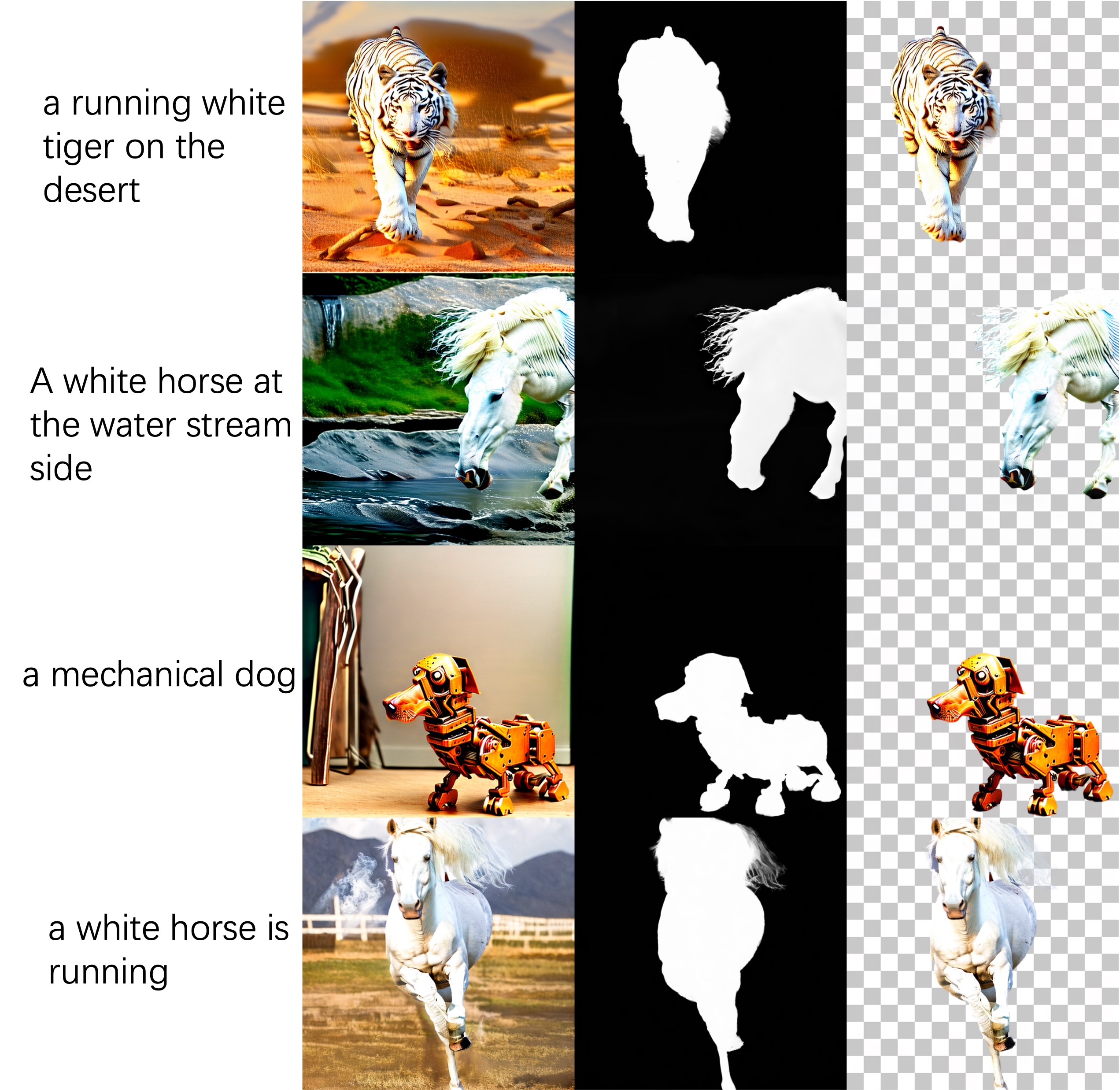}
\end{center}
\caption{More jointly generated results from Zippo.}
\label{fig:am2k2}
\end{figure*}

\begin{figure*}[ht!]
\begin{center}
\includegraphics[width=1.0\linewidth]{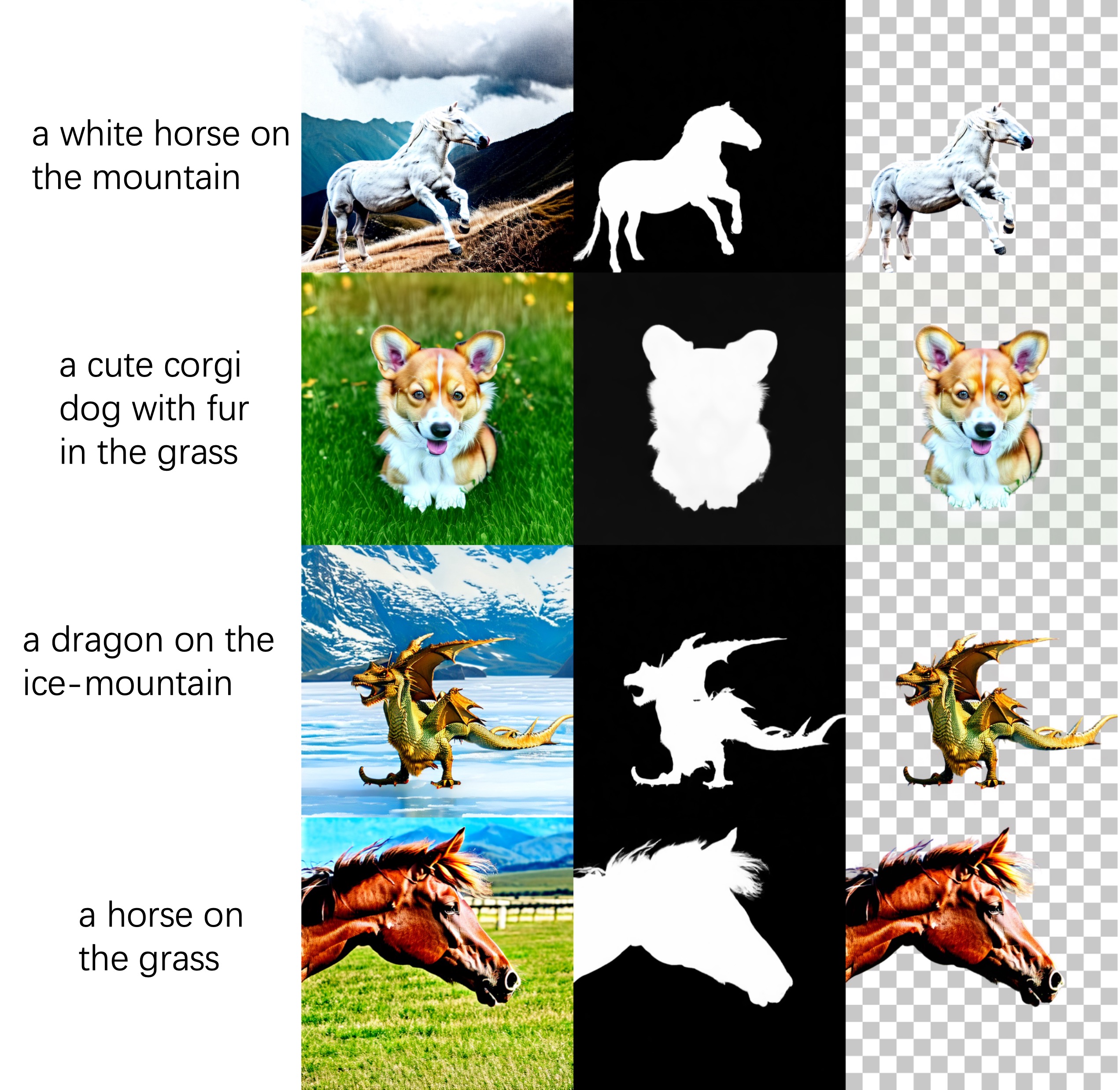}
\end{center}
\caption{More jointly generated results from Zippo.}
\label{fig:am2k3}
\end{figure*}

\begin{figure*}[ht!]
\begin{center}
\includegraphics[width=1.0\linewidth]{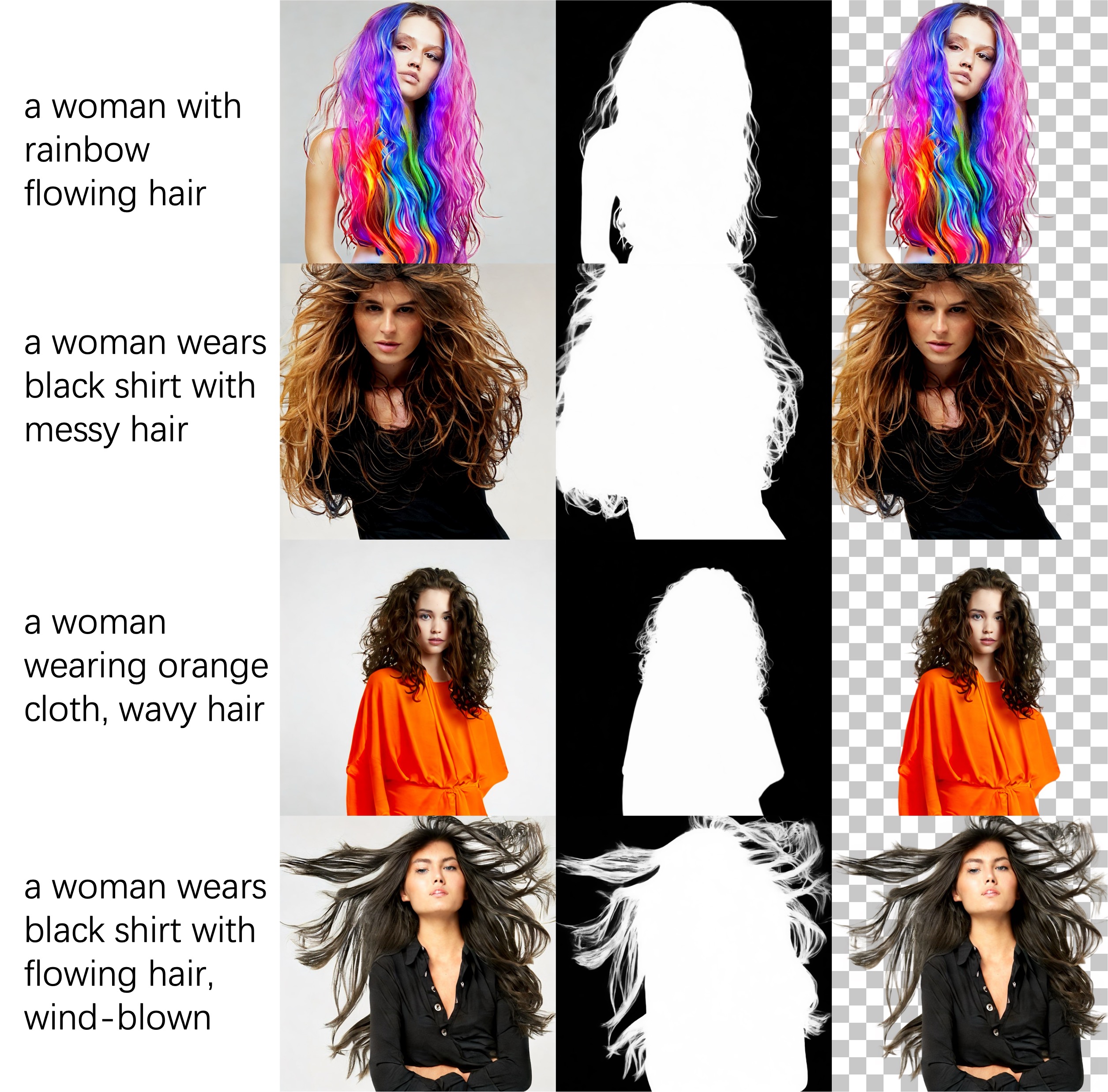}
\end{center}
\caption{More jointly generated results from Zippo.}
\label{fig:viton1}
\end{figure*}

\begin{figure*}[ht!]
\begin{center}
\includegraphics[width=1.0\linewidth]{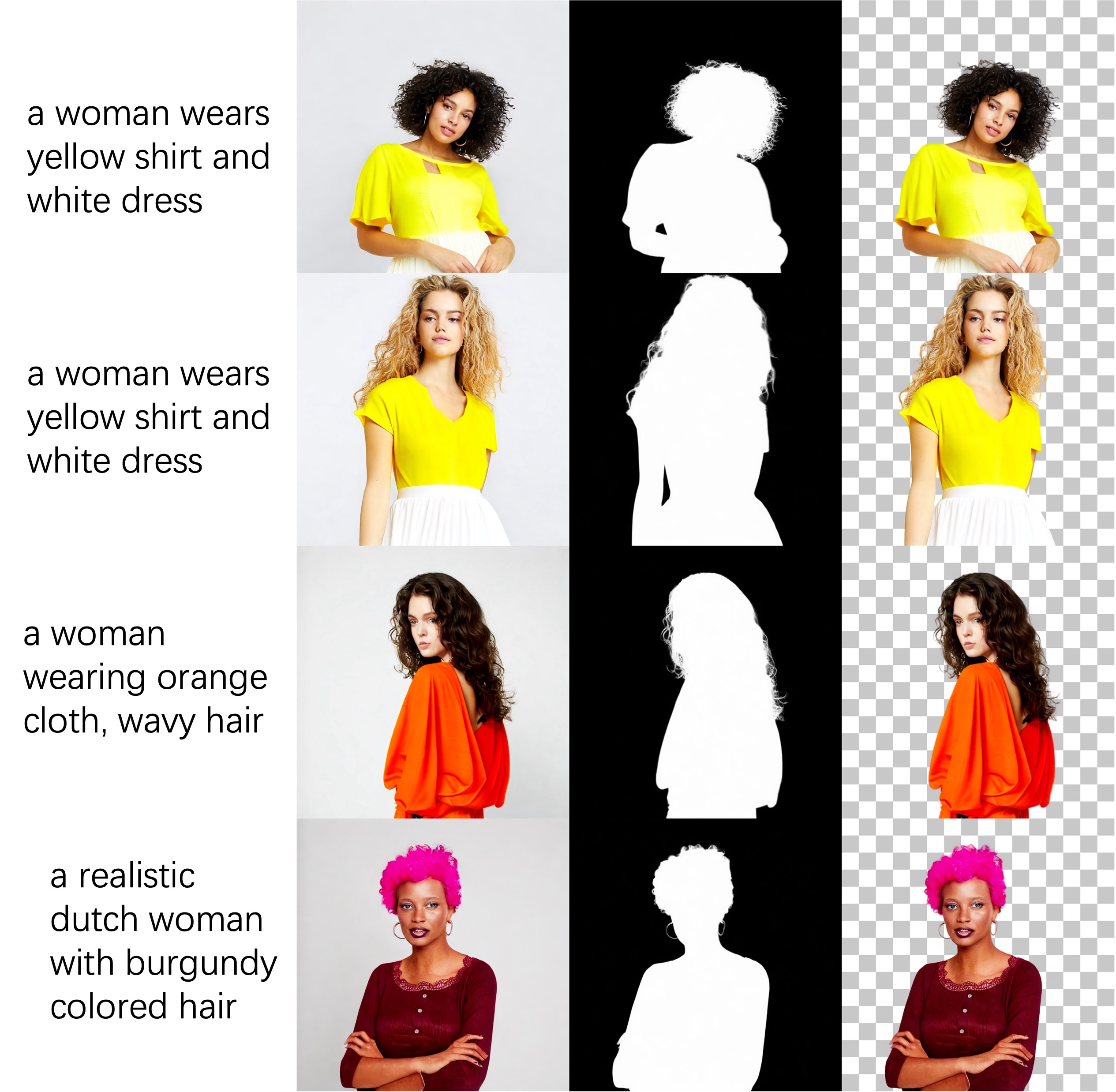}
\end{center}
\caption{More jointly generated results from Zippo.}
\label{fig:viton2}
\end{figure*}

\begin{figure*}[ht!]
\begin{center}
\includegraphics[width=1.0\linewidth]{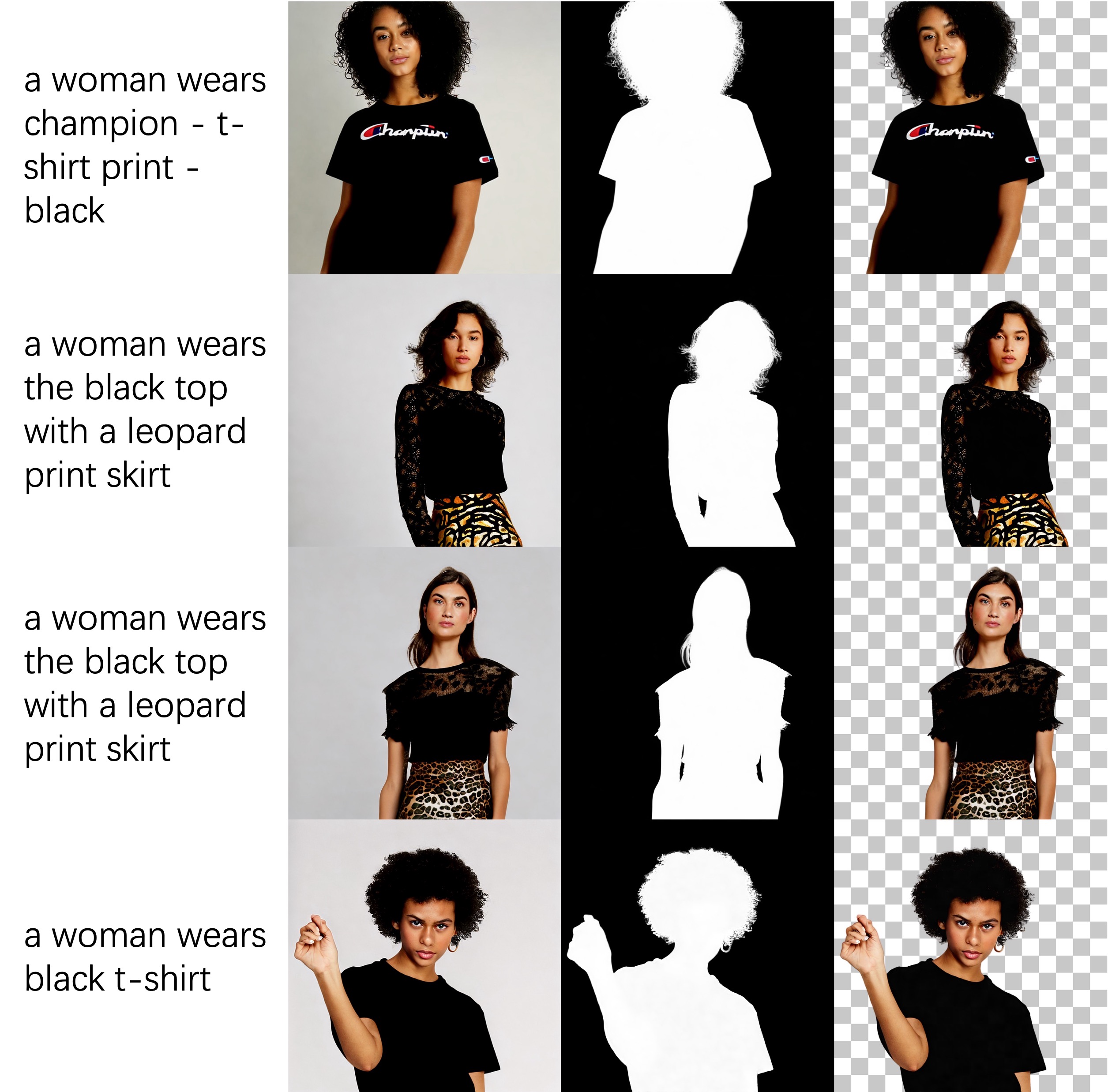}
\end{center}
\caption{More jointly generated results from Zippo.}
\label{fig:viton3}
\end{figure*}

\clearpage

\bibliographystyle{splncs04}
\bibliography{main}
\end{document}